\documentclass[runningheads]{llncs}
\usepackage[T1]{fontenc}
\usepackage{graphicx}
\usepackage{graphicx}
\usepackage{tcolorbox}
\usepackage{amsmath} 
\usepackage{amssymb}
\usepackage{subcaption}
\usepackage{hyperref}
\usepackage[ruled,vlined]{algorithm2e}

\begin{document}
\title{Benchmarking Uncertainty Metrics for LLM Prompt Optimization}
%

\author{Pei-Fu Guo\inst{1,2} \and
Yun-Da Tsai\inst{1,3} \and
Shou-De Lin\inst{1,4}}

\authorrunning{Guo. et al.}

\institute{National Taiwan University \and
\email{r12922217@csie.ntu.edu.tw}\\
\and \email{f08946007@csie.ntu.edu.tw} \\
\and \email{sdlin@csie.ntu.edu.tw}}


\maketitle              
\begin{abstract}
Prompting methods for LLMs like Chain of Thought (CoT) and Tree of Thought (ToT) improve reasoning through single- or multi-step processes. These methods can be enhanced by search algorithms, such as Monte Carlo Tree Search (MCTS) and Bandit methods, which rely on accurate uncertainty estimation. However, current uncertainty metrics for text generation—based on token-level likelihoods or verbalized confidence—primarily focus on output variability and do not align with the needs of optimization.

In this work, we first highlight the different requirements for uncertainty metrics in prompt optimization versus text generation. We outline four key uncertainties—Answer, Correctness, Aleatoric, and Epistemic—that are beneficial for prompt optimization and introduce a novel benchmarking pipeline to evaluate how well current NLG uncertainty metrics estimate these target uncertainties. 

Our experiments using GPT-3.5-Turbo and Meta-Llama-3.1-8B-Instruct on two reasoning datasets reveal a significant limitation in current uncertainty metrics, as they predominantly capture Answer Uncertainty but fail to effectively measure other types of uncertainty. This gap highlights the need for a broader range of optimization-aware uncertainty estimators to effectively guide search in prompt optimization tasks with different objectives. Our code and data are available at \href{https://github.com/0Frett/PO-Uncertainty-Benchmarking}{github link}.

\end{abstract}
\section{Introduction}
Prompting methods for large language models (LLMs) have gained significant attention for their ability to enhance reasoning capabilities through multi-step processes, such as Chain of Thought (CoT)\cite{wei2022chain}, Tree of Thought(ToT)\cite{yao2024tree}, and ReAct\cite{yao2022react}. These approaches can be extended by incorporating search algorithms to optimize prompts, utilizing techniques like Monte Carlo Tree Search (LATS, STaR)\cite{zhou2023language,zelikman2022star}, bandit algorithms (LongPO)\cite{hsieh2023automatic}, and gradient-style search (OPRO)\cite{Yang2023LargeLM}. A key element in these search and optimization algorithms is uncertainty estimation, which is vital for guiding decisions, balancing exploration and exploitation, and improving algorithm efficiency. Uncertainty estimation techniques, such as those used in Bandit algorithms or Bayesian optimization, can dynamically adjust learning rates or hyperparameters. In combinatorial optimization (e.g., genetic algorithms or simulated annealing), uncertainty estimation informs heuristic decisions like mutation rates or temperature adjustments. Hence, developing robust methods to quantify uncertainty in LLMs, particularly for prompt optimization, is essential.

Previous approaches to measuring uncertainty in LLMs primarily rely on token-level or sentence-level generation likelihoods, often represented by metrics like token disparity probability~\cite{wang2024chain}, predictive entropy~\cite{npe_lnpe} and reciprocal of perplexity~\cite{chen1998evaluation}. These techniques have been used for bias calibration~\cite{zhou2023batch}, controllable decoding~\cite{zhu2024hot}, and LLM planning~\cite{ren2023robots}. However, we argue that such token-level or sentence-level uncertainty measurements are more indicative of model output confidence or output diversity, which may not align with the needs of prompt optimization tasks. In these contexts, uncertainty estimation should guide the search process itself. For example, in tree-based reasoning, the uncertainty at each node should help steer the search direction in line with the search objectives rather than simply reflecting model confidence or output variability.

\begin{figure*}[t!]
\centering
\begin{subfigure}[t]{0.48\textwidth}
    \centering
    \includegraphics[width=\textwidth]{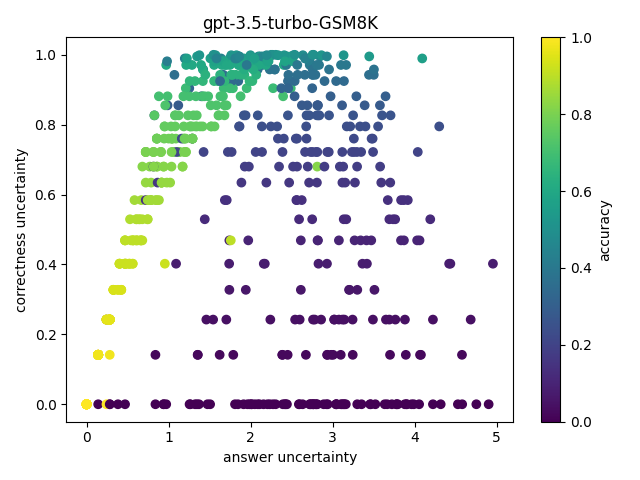}
    \caption{Relationship between correctness uncertainty, answer uncertainty and response accuracy. Each spot is a node within reasoning traces. See section ~\ref{sec:pipeline}.}
    \label{fig:au-cu}
\end{subfigure}
\hfill
\begin{subfigure}[t]{0.48\textwidth}
    \centering
    \includegraphics[width=\textwidth]{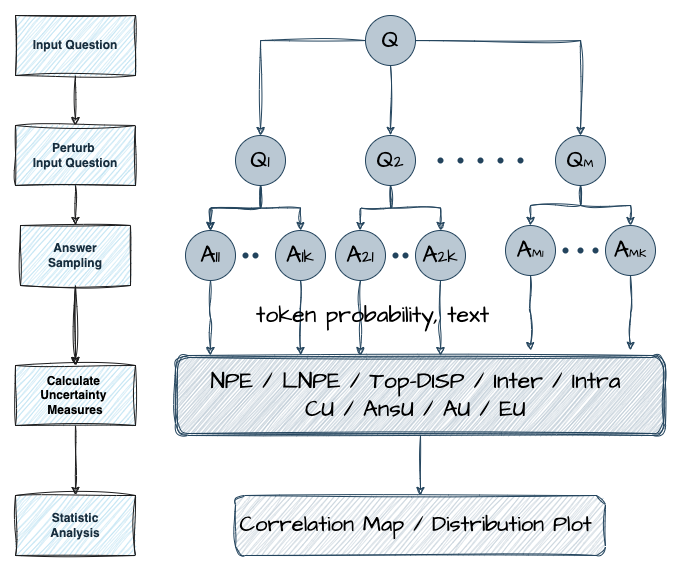}
    \caption{Overview of our benchmarking pipeline. Detail of each step is shown at Algorithm ~\ref{alg:benchmarking_pipeline}.}
    \label{fig:workflow}
\end{subfigure}
\label{fig:combined}
\end{figure*}

Figure\ref{fig:au-cu} justify our hypothesis and illustrates the relationship between LLM correctness uncertainty, answer uncertainty (see section ~\ref{sec:uncertainty_type}), and response accuracy in the GSM8K dataset. In prompt optimization tasks focused on searching correct answer, a reliable uncertainty metric targeting this objective should exhibits 50\% response accuracy(correct/wrong) when its value is at its highest. This objective conflicts with answer uncertainty, which is designed to measure diversity of responses but may reflect an incorrect majority answer.

In this work, we first highlight the differing requirements for uncertainty metrics in prompt optimization versus text generation. We outline four key uncertainties that are beneficial to prompt optimization—Answer, Correctness, Aleatoric, Epistemic and propose a novel benchmarking pipeline designed to evaluate the effectiveness of current NLG uncertainty metrics in prompt optimization setting. By performing extensive sampling on LLMs, our pipeline construct large, tree-structured reasoning traces from model outputs. Once these traces are built, we can compute accurate estimation of the uncertainties, which can serve as ground truth values for comparison with metric predictions. Our evaluation shows that current uncertainty metrics mainly capture Answer Uncertainty and fail to measure other uncertainty types, emphasizing the need for more diverse, optimization-aware estimators to guide prompt optimization for different objectives.

\section{Different Uncertainties for Prompt Optimization}
\label{sec:uncertainty_type}
Uncertainty can arise from various aspects and express in different forms. For instance, uncertainty about the input question might reflect ambiguity or lack of clarity in the prompt itself, whereas uncertainty about the output answer concerns the model fidelity or diversity of the responses generated. Each type of uncertainty represents different dimensions of the problem and can guide the optimization process in unique ways. In this section, we outline four types of uncertainty, each playing a distinct role and offering unique benefits for prompt optimization across tasks.

\textbf{Answer Uncertainty (AnsU)} reflects the model's confidence and the diversity of possible answers. AnsU describes how consistently the model produces the same answer after repeated sampling, but it does not guarantee correctness. If the model lacks necessary knowledge to answer the given question, it is reasonable for the output answers to be incorrect, even if low AnsU is observed after repeated sampling.

AnsU is measured as the entropy of the output answer distribution.
\vspace{-2mm}
\[
AnsU(x) = - \sum_{i} p(y_i | x) \log p(y_i | x)
\]
where \(p(y_i | x)\) is the probability of output answer \(y_i\) obtain from \(\{\mathbf{x}_j\}\) given the input \(x\). AnsU can guide prompt algorithms to explore a richer solution space, increasing the variability of generated content. This can be particularly beneficial for tasks like creative writing, idea generation and open-ended problem-solving, where diversity and originality are highly valued.

\textbf{Correctness Uncertainty (CU)} provides insights into the likelihood of answer correctness. For example, in a medical diagnosis system, high CU indicates that the model's prediction may be unreliable, suggesting the need for additional verification or consultation. Note that this differs from AnsU; CU is directly related to the accuracy of the diagnosis. When CU is low, the model's predictions are less likely to include a majority of false positives.

CU is calculated as the entropy of the output correctness distribution.
\vspace{-2mm}
\[
CU(x) = - \sum_{i} p(c_i | x) \log p(c_i | x)
\]
where \(p(c_i | x)\) represents the probability of correctness \(c_i\) (whether an answer is correct or incorrect) obtain from \(\{\mathbf{x}_j\}\) given the input \(x\). CU can guide prompt algorithms to effectively narrow down the solution space and acquire the correct answer. In question-answering tasks with a ground truth answer, CU helps the model focus on more accurate responses, reducing the likelihood of incorrect or irrelevant answers.

\textbf{Aleatoric Uncertainty (AU)} and \textbf{Epistemic Uncertainty (EU)} are two distinct sources of model uncertainty. AU arises from the inherent vagueness or noise in the data itself. For example, if asked "What time is the meeting?", even if the LLM has access to a detailed schedule, it may still be uncertain because the question doesn't specify which meeting is being referred to, making it inherently ambiguous. EU, on the other hand, originates from the model's limitations and is related to its knowledge and understanding. This uncertainty can be reduced by training the model with more data or refining its algorithms. For instance, a language model trained on a limited dataset might show high EU in underrepresented domains or languages.

AU and EU are calculated using the Deep-Ensemble-Decomposition method \cite{decompose}, where total model uncertainty is the sum of AU and EU. In our context, \(\theta\), which represents model parameters in the original paper, corresponds to the perturbed question in our setting. EU captures the disagreement between different perturbations, measured by the mutual information \(I(Y ; \theta | X)\). AU reflects the inherent data noise and is represented as \(\mathbb{E}_{q(\theta|D)}[H(q(Y | X, \theta))]\), where the expectation is taken over the perturbed questions \(\theta\). 

The total model uncertainty is expressed as :
\small
\[
H(q(Y | X)) = \underbrace{I(Y ; \theta | X)}_{\text{EU}} + \underbrace{\mathbb{E}_{q(\theta|D)}[H(q(Y | X, \theta))]}_{\text{AU}}
\]
\normalsize
The advantage of AU and EU is that they explain the underlying causes of uncertainty(due to data noise or model limitations). This understanding enables more targeted improvements during prompt optimization, such as rephrasing the question or providing additional few-shot examples.

\section{Current NLG Uncertainty Metrics}
\label{sec:metric}
Current NLG uncertainty metrics measure the uncertainty in a model's output and aim to improve correctness by favoring answers with lower uncertainty, under the assumption that more model confident answers are more likely to be correct. In this section, we outline four commonly used black-box metrics, primarily based on decoding probabilities and verbalized confidence. We further evaluate their effectiveness in quantifying different uncertainties through our benchmarking pipeline in section~\ref{sec:pipeline}.

\textbf{Normalized Predictive Entropy (NPE)}\cite{npe_lnpe} measures the uncertainty of generated text by calculating the average entropy of possible output sequences given a input context \(\mathbf{x}\). 

\vspace{-6mm}
\begin{align*}
\text{NPE}(\mathbf{x})&= \frac{1}{N} \sum_{n} \sum_{i} \ln p(s_i | s_{<i})
\end{align*}
where \(N\) is the number of generations and \(s_i\) is the \(i\)-th token of sentence $\mathbf{s}$.
\vspace{1mm}

\textbf{Length-Normalized Predictive Entropy (LNPE)}\cite{npe_lnpe} adjusts for sentence length by normalizing the entropy with the number of tokens. This ensures fair comparison across different sentence lengths \(S_{n}\) .
\vspace{-2mm}
\begin{align*}
\text{LNPE}(\mathbf{x}) &= \frac{1}{N} \sum_{n} \left(\frac{1}{S_{n}}\right) \sum_{i} \ln p(s_i | s_{<i})
\end{align*}

\vspace{-2mm}
\textbf{TopK-Token Disparity (Top-DISP)} is based on the concept introduced by \cite{disparity}, which suggests that a larger difference between the top-1 and top-2 token probabilities correlates with higher confidence in the model answer. The metric calculates the average difference in probability between the top-1 and top-2 tokens for each token within the output sequence and further averages them across multiple outputs.

\small
\vspace{-2mm}
\begin{align*}
\text{Top-DISP}(\mathbf{x})
    &= -\frac{1}{N} \sum_{n} \left(\frac{1}{S_{n}}\right) \sum_{i} \left| \ln \frac{p(s_{i,\text{top1}} | s_{<i})}{p(s_{i,\text{top2}} | s_{<i})} \right|
\end{align*}
\normalsize

\normalsize
\textbf{Intra-Sample Similarity (Intra)}\cite{spuq} computes the average of the uncertainties discerned individually for each sample output. Following the approach used in SPUQ, we utilize verbalized uncertainty method \cite{verb} to obtain the uncertainty articulated by the LLM for each perturbed input and output pair \(c(x_i, y_i)\). 
\vspace{-2mm}
\begin{align*}
\text{Intra}(\mathbf{x}) &= -\frac{\sum_{i=0}^{k} c(x_i, y_i)}{k + 1}
\end{align*}

\section{Benchmarking Pipeline}
\label{sec:pipeline}
A reliable uncertainty metric should serve as an accurate estimator of its target uncertainty. In this study, we introduce a novel benchmarking pipeline designed to assess the effectiveness of uncertainty metrics in estimating target uncertainties within the context of prompt optimization. Our pipeline focuses on evaluating metrics in tree-structured reasoning traces, which represent a predominant approach in current prompting algorithms. The following sections outline the conceptual foundation and steps of our pipeline.

\subsection{Design Concept}
To evaluate how well a metric quantifies its target uncertainty in prompt optimization, we first need to establish the ground truth values for the target uncertainty at each reasoning step. This requires constructing a comprehensive set of reasoning traces for input questions.

We generate large tree-structured reasoning traces for each question by perturbing input prompts and sampling outputs multiple times in each node of traces. This servies two key purposes. First, these traces align with prompt optimization algorithms, as they emulate how such algorithms explore possible reasoning paths, making them highly relevant for evaluating metrics in realistic scenarios. Second, they enable a thorough exploration of the solution space by considering diverse reasoning paths. This approach allows the tree-structured traces to approximate the complete solution space, facilitating robust estimation of target uncertainty ground truth values using Monte Carlo methods.

After obtaining the ground truth values, we calculate the metric estimates at each reasoning node based on the formula in section ~\ref{sec:metric}. We then evaluate the alignment between uncertainty metrics and ground truth values using statistical methods. This involves measuring the correlation, bias, and variance of the metric estimates relative to the ground truth. Figure~\ref{fig:workflow} shows an overview of our benchmarking pipeline.

\subsection{Detailed Workflow}
\label{sec:workflow}

Given a dataset of questions, our pipeline builds a reasoning tree for each question, ultimately producing a large number of \((U_{\text{metric}}, U_{\text{true}})\) pairs. Algorithm~\ref{alg:benchmarking_pipeline} outlines the process in detail. Increasing \(M\) and \(K\) improves the approximation to the underlying solution space, yielding more accurate ground truth values and consequently improving the quality of the evaluation.

\begin{algorithm}[h!]
\caption{Benchmarking Pipeline Workflow}
\label{alg:benchmarking_pipeline}
\ForEach{input question \(Q\)}{
    Initialize a reasoning tree with a root node containing \(Q\)\;
    
    \While{the reasoning tree is not fully constructed}{
        \ForEach{node in the tree that has not yet terminated}{
            \textbf{Step 1: Input Perturbation} \\
            Generate \(M\) rephrased inputs \(\{x_j\}_{j=1}^M\) from the current node’s input \(\mathbf{x}\)\;

            \textbf{Step 2: Random Sampling} \\
            For each rephrased input \(\mathbf{x}_j\), sample \(K\) responses \(\{y_{jk}\}_{k=1}^K\)\;
        }
        Expand the tree by adding child nodes using the newly generated responses\;
    }

    \textbf{Step 3: Ground Truth Uncertainty Calculation} \\
    For each node, calculate the ground truth uncertainty using the answers in its subtree’s leaves, as described in Section~\ref{sec:uncertainty_type}\;

    \textbf{Step 4: Uncertainty Metric Calculation} \\
    For each node, compute the estimated uncertainty metric based on its own input and output, following the formulas in Section~\ref{sec:metric}\;
}

\textbf{Step 5: Statistical Analysis} \\
Collect all \((U_{\text{metric}}, U_{\text{true}})\) pairs from every node across all trees.\;
Compute \(\text{Corr}(U_{\text{metric}}, U_{\text{true}})\) and visualize these relationships to assess how well the estimated metrics align with the ground truth.\;
\end{algorithm}

\section{Experiments}
In this section, we use our benchmarking pipeline to evaluate the uncertainty metrics introduced in Section~\ref{sec:metric} against the target uncertainty defined in Section~\ref{sec:uncertainty_type}. Our evaluation focuses on the correlation map (see ~\ref{fig:corrs}) and visualization plots (see ~\ref{app:scatter}) between uncertainty metrics and ground truth uncertainty values. By analyzing these relationships, we can determine which metrics serve as better estimators of the four target uncertainties, guiding prompt optimization more effectively. Prompt templates are shown in ~\ref{app:prompt}.

\subsection{Dataset and LLMs}
We conduct experiments on two reasoning datasets: GSM8K~\cite{cobbe2021training} and StrategyQA~\cite{geva2021did}, which involve solving math problems and complex strategic reasoning, respectively. We selected these datasets because they provide exact ground truth answers and encompass diverse types of knowledge and problem-solving tasks. While both are reasoning datasets, GSM8K features an infinite range of possible answers, whereas StrategyQA is constrained to binary answers: "true" or "false". We use two large language models, GPT-3.5-Turbo\cite{openai} and Meta-Llama-3.1-8B-Instruct, to showcase that our benchmarking pipeline is suitable for both commercial and open-source LLMs of varying sizes.

\subsection{Results and Analysis}
\begin{figure*}[t!]
\vspace{-1cm}
  \centering
  \begin{subfigure}[b]{0.45\textwidth}
    \centering
    \includegraphics[width=\textwidth]{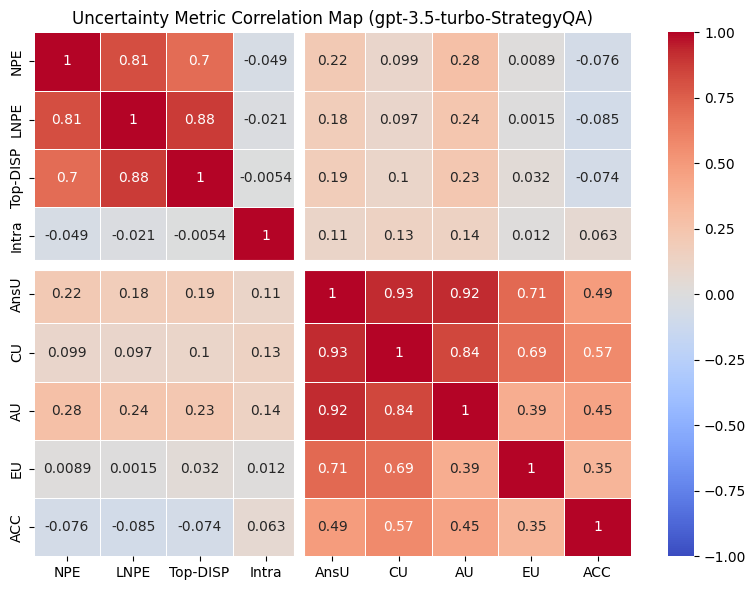}
    \label{fig:strategy_gpt_corr_map}
  \end{subfigure}
  \hspace{0.01\textwidth} 
  \begin{subfigure}[b]{0.45\textwidth}
    \centering
    \includegraphics[width=\textwidth]{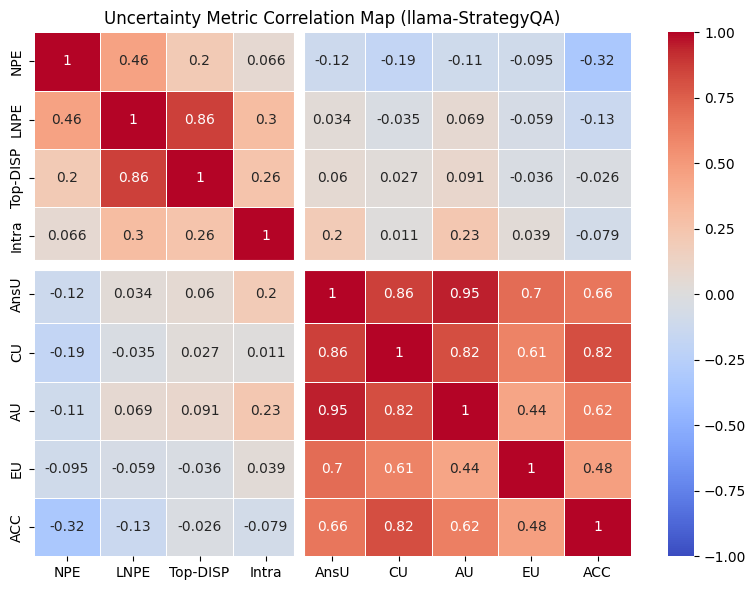}
    \label{fig:strategy_llama_corr_map}
  \end{subfigure}

  \vspace{0.02in} 

  \begin{subfigure}[b]{0.45\textwidth}
    \centering
    \includegraphics[width=\textwidth]{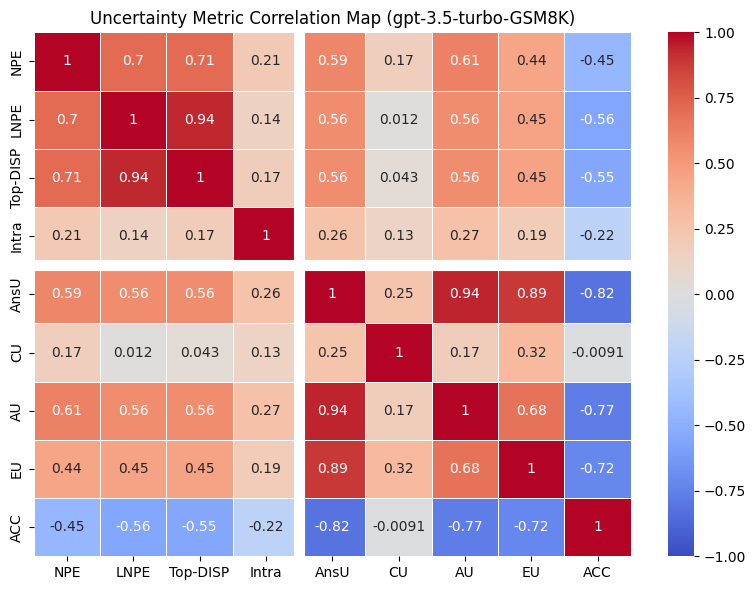}
    \label{fig:gsm8k_gpt_corr_map}
  \end{subfigure}
  \hspace{0.01\textwidth}
  \begin{subfigure}[b]{0.45\textwidth}
    \centering
    \includegraphics[width=\textwidth]{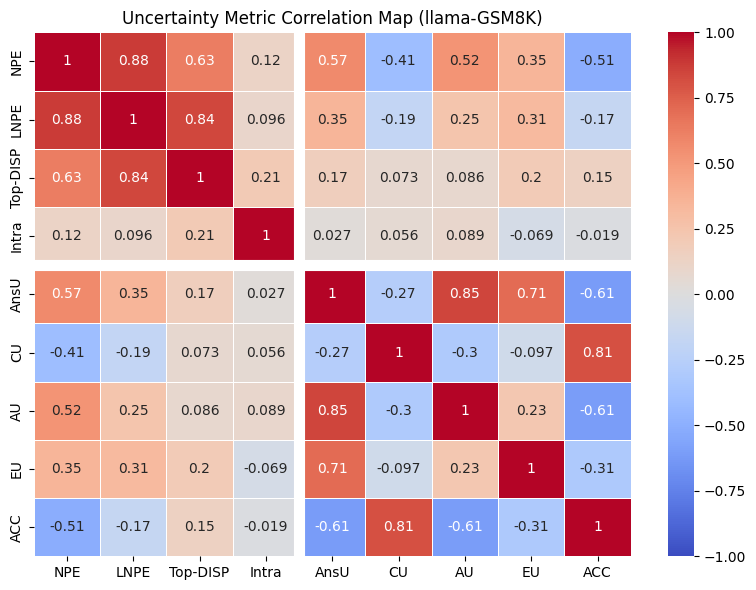}
    \label{fig:gsm8k_llama_corr_map}
  \end{subfigure}

  \caption{Correlation Maps of uncertainty metrics and target uncertainty on different datasets and models.}
  \label{fig:corrs}
\end{figure*}

Figure~\ref{fig:corrs} presents correlation maps illustrating the relationships between uncertainty metrics and target uncertainties. Each map is divided into three sections: the upper-left shows correlations among the uncertainty metrics, the upper-right shows how well each metric aligns with the target uncertainties, and the lower-right shows correlations among the target uncertainties themselves. A strong correlation between a metric and a target uncertainty indicates that the metric serves as an effective estimator, which can be used to guide search in prompting algorithms. Below, we summarize key findings:

\begin{enumerate}
    \item \textbf{Current metrics estimate AnsU well but struggle with CU.} As shown in Figure~\ref{fig:corrs}, the correlation maps demonstrate that evaluated metrics are more correlated with AnsU and AU. As for CU, we can see zero or negative correlation in all the maps. This implies most uncertainty metrics predominantly capture uncertainty of answer diversity and model confidence, but fail to estimate the correctness, which is important in prompting algorithms targeting answer correctness.
    
    \item \textbf{Uncertainty metrics show strong inter-correlation.} The upper-left section of each map shows that token-likelihood-based metrics (NPE, LNPE, Top-DISP) are highly correlated, while INTRA, a verbalized confidence metric, exhibits little correlation with them. This high correlation indicates that these estimators capture similar uncertainties, highlighting a lack of diversity in metrics for other uncertainty types, such as CU.
    
    \item \textbf{AnsU's weak link to CU reveals fundamental differences of the two.} The lower-right section of the correlation maps shows a strong relationship between AnsU, AU, and EU, suggesting that these uncertainties can potentially share common estimators in prompting algorithms. However, CU displays a much weaker or even negative correlation with AnsU, especially in the GSM8K dataset when solution space are really large. This highlights a key distinction: AnsU reflects answer diversity, CU measures answer correctness. Metrics designed to estimate AnsU are not effective for capturing CU. Therefore, tailored estimators for correctness uncertainty are needed to better support tasks focused on accuracy.
    
\end{enumerate}

\section{Conclusion}
In this work, we emphasize the different requirements for uncertainty metrics in prompt optimization versus text generation. We propose a novel benchmarking pipeline to assess how well current NLG uncertainty metrics estimate various types of uncertainty in prompt optimization settings. Through experiments on two datasets and multiple LLMs, we find that most NLG uncertainty metrics primarily capture uncertainty related to answer diversity or model confidence, but fail to estimate uncertainty related to correctness, which is an essential factor for prompting algorithms targeting accurate answers. This gap underscores the need for optimization-aware uncertainty metrics that can better guide prompt optimization in LLMs.

%
%
\bibliographystyle{splncs04} 
\bibliography{custom} 

\appendix

\section{Prompt Templates}
\label{app:prompt}

\begin{figure}[h]
\begin{tcolorbox}[width=1.0\linewidth, halign=left, colframe=black, colback=white, boxsep=0.01mm, arc=1.5mm, left=2mm, right=2mm, boxrule=0.5pt]\footnotesize

You will receive a question and your goal is to generate a new version of it that convey the same meaning as the original.\\
Q1.Original Question: Would a dog respond to bell before Grey seal? \\
New-Version: Would a dog react to a bell sooner than a grey seal? \\
Q2.Original Question: The perimeter of a rectangle is the sum of all its sides.\\
New-Version: A rectangle's perimeter is obtained by summing the lengths of its sides.\\
\textbf{Q3. Original Question:} <Question>\\
\textbf{New-Version:}

\end{tcolorbox}
\caption{Example prompt for question perturbation.}
\label{fig:prompt-objfunc1}
\end{figure}

\begin{figure}[h]
\begin{tcolorbox}[width=1.0\linewidth, halign=left, colframe=black, colback=white, boxsep=0.01mm, arc=1.5mm, left=2mm, right=2mm, boxrule=0.5pt]\footnotesize

Q: Was ethanol beneficial to Jack Kerouac's health?\\
A: Jack Kerouac died from internal bleeding due to long-term alcohol abuse. Thus, ethanol was not beneficial to Jack Kerouac's health. So the answer is no.\\
Q: If Goofy were a pet, would he need heartworm prevention?\\
A: Goofy is a dog, and dogs require regular heartworm prevention. Thus, if Goofy were a pet, he would need heartworm prevention. So the answer is yes.\\
\textbf{Q :} <Question>\\ 
\textbf{A :}\\

\end{tcolorbox}
\caption{Example prompt for StrategyQA. We random pick 4 few shot samples from pool.}
\label{fig:prompt-objfunc2}
\end{figure}

\begin{figure}[h]
\begin{tcolorbox}[width=1.0\linewidth, halign=left, colframe=black, colback=white, boxsep=0.01mm, arc=1.5mm, left=2mm, right=2mm, boxrule=0.5pt]\footnotesize

Question 1: Mark has a garden with flowers. He planted plants of three different colors in it. Ten of them are yellow, and there are 80\% more of those in purple. There are only 25\% as many green flowers as there are yellow and purple flowers. How many flowers does Mark have in his garden?\\
Answer : There are 80\% more purple flowers than yellow flowers, so there are 10 * 1.8 = 18 purple flowers. There are 10 yellow flowers and 18 purple flowers, so there are 10 + 18 = 28 yellow and purple flowers. There are 25\% as many green flowers as there are yellow and purple flowers, so there are 28 * 0.25 = 7 green flowers. Mark has 10 yellow flowers, 18 purple flowers, and 7 green flowers, so he has 10 + 18 + 7 = 35 flowers in his garden. The answer to the question is 35.\\
Question 2: Albert is wondering how much pizza he can eat in one day. He buys 2 large pizzas and 2 small pizzas. A large pizza has 16 slices and a small pizza has 8 slices. If he eats it all, how many pieces does he eat that day?\\
Answer : He buys 2 large pizzas, so he has 2 * 16 = 32 slices. He buys 2 small pizzas, so he has 2 * 8 = 16 slices. There are 32 slices from the large pizzas and 16 slices from the small pizzas, so he eats 32 + 16 = 48 pieces that day. The answer to the question is 48.\\
\textbf{Question 3:} <Question>\\
\textbf{Answer :}
\end{tcolorbox}
\caption{Example prompt for GSM8K. We random pick 4 few shot samples from pool.}
\label{fig:prompt-objfunc3}
\end{figure}

\clearpage
\section{Additional Results}
\label{app:scatter}
\begin{figure*}[h!]
  \centering
  \begin{subfigure}[b]{0.24\textwidth}
    \centering
    \includegraphics[width=\textwidth]{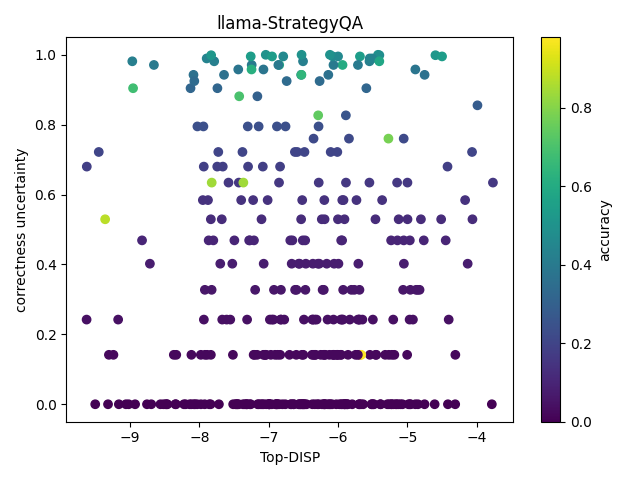}
    \label{fig:strategy_llama_disp}
  \end{subfigure}
  \hfill
  \begin{subfigure}[b]{0.24\textwidth}
    \centering
    \includegraphics[width=\textwidth]{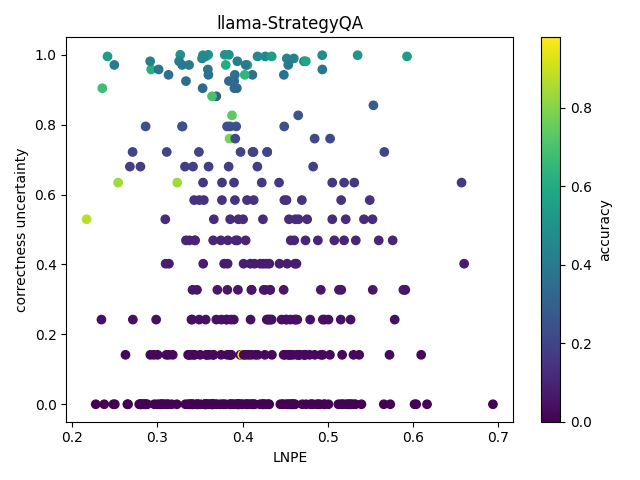}
    \label{fig:strategy_llama_lnpe}
  \end{subfigure}
  \hfill
  \begin{subfigure}[b]{0.24\textwidth}
    \centering
    \includegraphics[width=\textwidth]{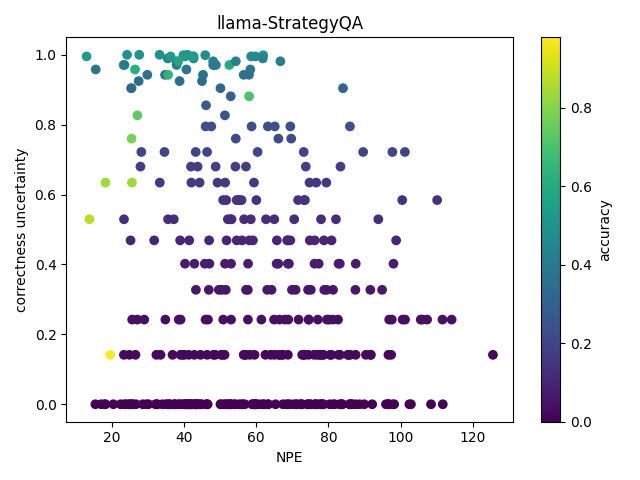}
    \label{fig:strategy_llama_npe}
  \end{subfigure}
  \hfill
  \begin{subfigure}[b]{0.24\textwidth}
    \centering
    \includegraphics[width=\textwidth]{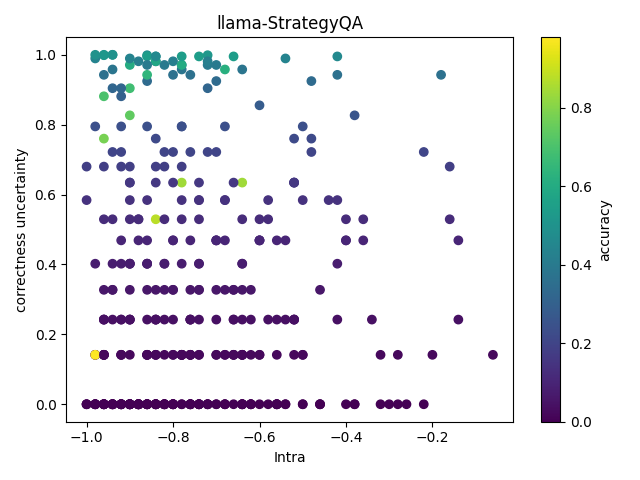}
    \label{fig:strategy_llama_intra}
  \end{subfigure}
  
  \vspace{0.2cm}

  \begin{subfigure}[b]{0.24\textwidth}
    \centering
    \includegraphics[width=\textwidth]{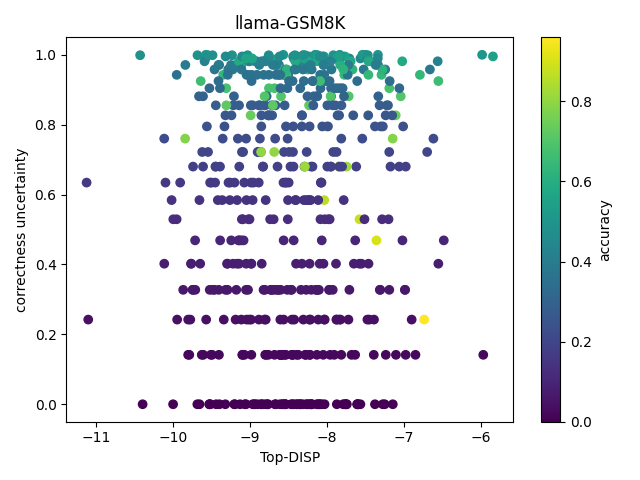}
    \label{fig:gsm8k_llama_disp}
  \end{subfigure}
  \hfill
  \begin{subfigure}[b]{0.24\textwidth}
    \centering
    \includegraphics[width=\textwidth]{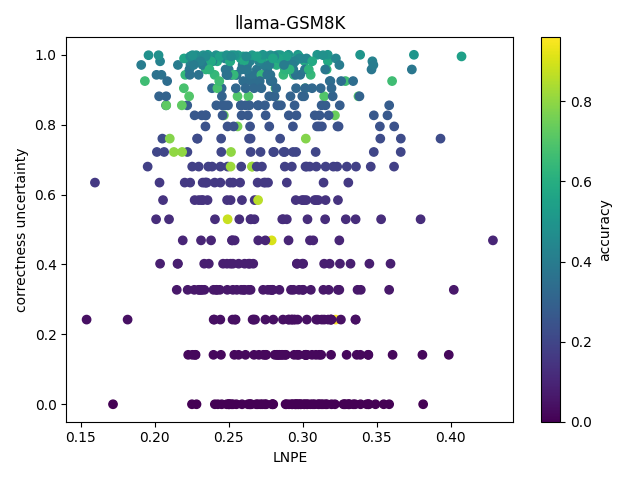}
    \label{fig:gsm8k_llama_lnpe}
  \end{subfigure}
  \hfill
  \begin{subfigure}[b]{0.24\textwidth}
    \centering
    \includegraphics[width=\textwidth]{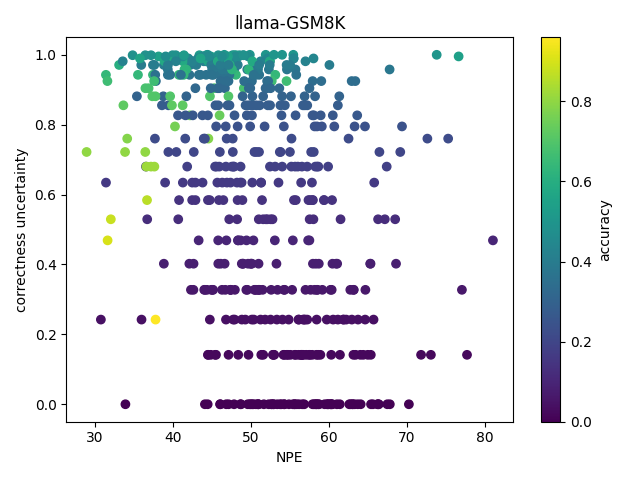}
    \label{fig:gsm8k_llama_npe}
  \end{subfigure}
  \hfill
  \begin{subfigure}[b]{0.24\textwidth}
    \centering
    \includegraphics[width=\textwidth]{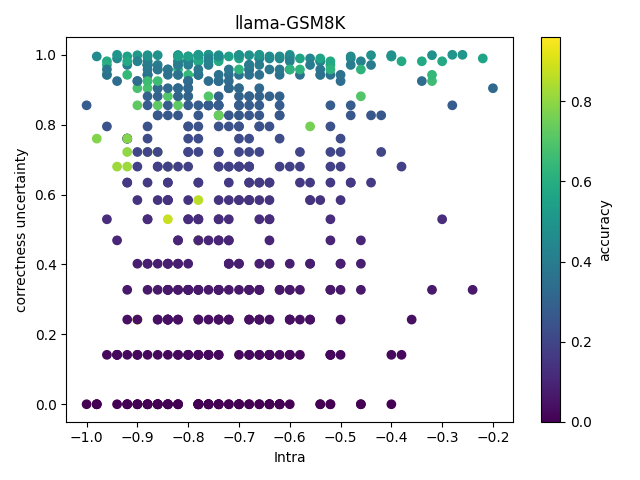}
    \label{fig:gsm8k_llama_intra}
  \end{subfigure}
  
  \vspace{0.2cm}

  \begin{subfigure}[b]{0.24\textwidth}
    \centering
    \includegraphics[width=\textwidth]{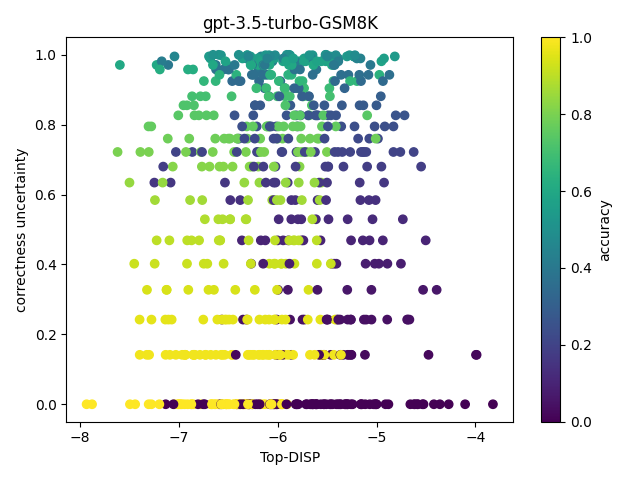}
    \label{fig:gsm8k_gpt_disp}
  \end{subfigure}
  \hfill
  \begin{subfigure}[b]{0.24\textwidth}
    \centering
    \includegraphics[width=\textwidth]{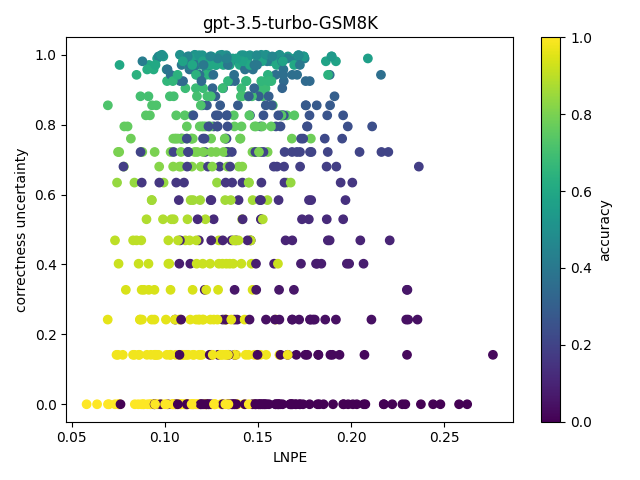}
    \label{fig:gsm8k_gpt_lnpe}
  \end{subfigure}
  \hfill
  \begin{subfigure}[b]{0.24\textwidth}
    \centering
    \includegraphics[width=\textwidth]{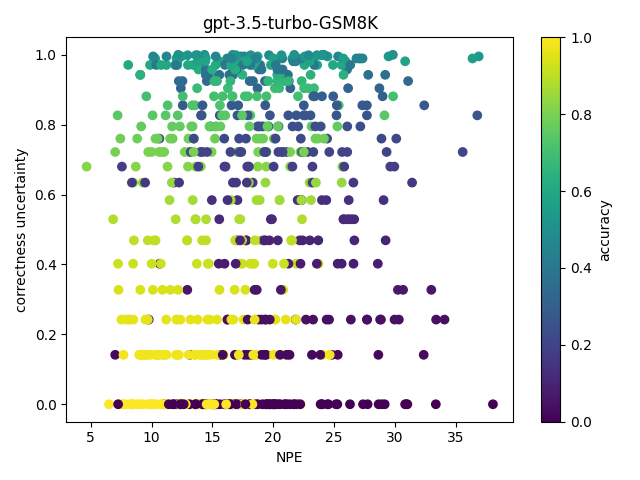}
    \label{fig:gsm8k_gpt_npe}
  \end{subfigure}
  \hfill
  \begin{subfigure}[b]{0.24\textwidth}
    \centering
    \includegraphics[width=\textwidth]{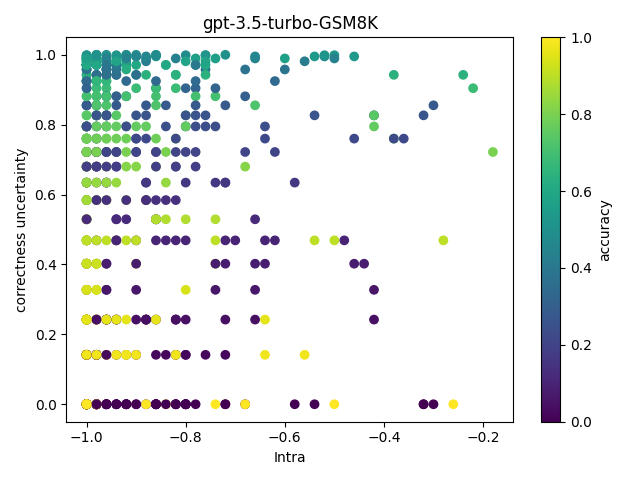}
    \label{fig:gsm8k_gpt_intra}
  \end{subfigure}
  
  \vspace{0.2cm}

  \begin{subfigure}[b]{0.24\textwidth}
    \centering
    \includegraphics[width=\textwidth]{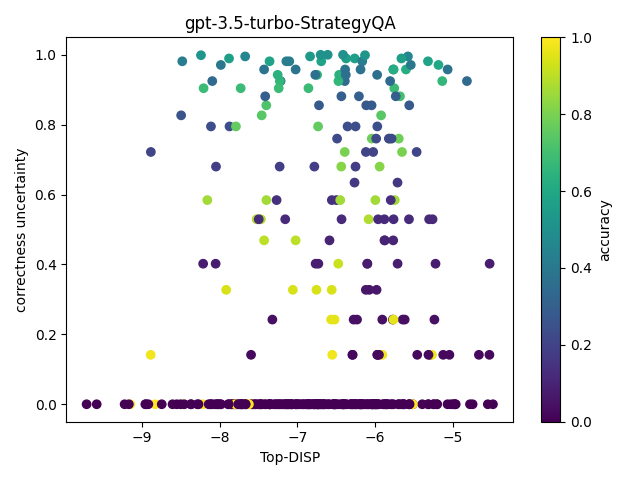}
    \label{fig:strategy_gpt_disp}
  \end{subfigure}
  \hfill
  \begin{subfigure}[b]{0.24\textwidth}
    \centering
    \includegraphics[width=\textwidth]{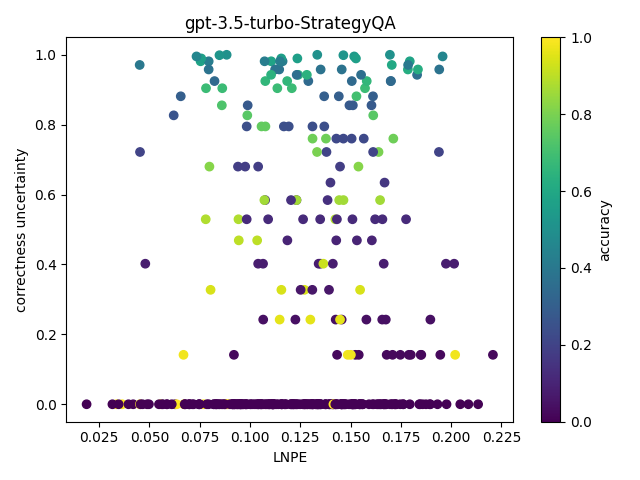}
    \label{fig:strategy_gpt_lpne}
  \end{subfigure}
  \hfill
  \begin{subfigure}[b]{0.24\textwidth}
    \centering
    \includegraphics[width=\textwidth]{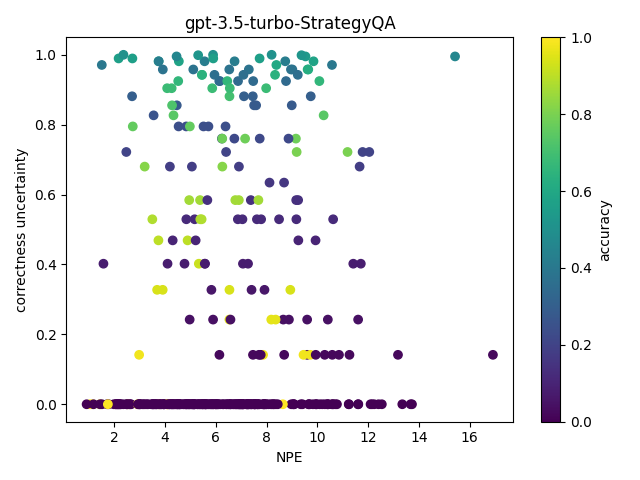}
    \label{fig:strategy_gpt_npe}
  \end{subfigure}
  \hfill
  \begin{subfigure}[b]{0.24\textwidth}
    \centering
    \includegraphics[width=\textwidth]{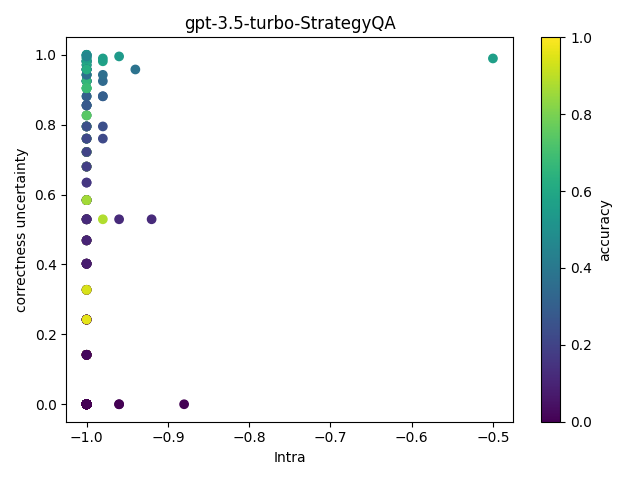}
    \label{fig:strategy_gpt_intra}
  \end{subfigure}

  \caption{Scatter plots show the evaluation results of metrics on Llama (StrategyQA and GSM8K) and GPT-3.5-Turbo (GSM8K and StrategyQA), with each point representing a reasoning node. The plots illustrate the relationship between CU, uncertainty metrics, and response accuracy. As shown, most metrics fail to estimate CU effectively, as there is no clear trend of higher metric values(x-axis) corresponding to higher correctness uncertainty(y-axis).}
  \label{fig:combined_scatter}
\end{figure*}

\end{document}